\documentclass[%
 aip,
 jmp,%
 amsmath,amssymb,
 reprint,
]{revtex4-2}

\usepackage{comment}
\usepackage{graphicx}
\usepackage{dcolumn}
\usepackage{bm}

\begin{document}
\title{Causal Atlases from Entropic Inference: Bayesian Networks beyond Optimal DAGs}

\author{Hazhir Aliahmadi}
\author{Irina Babayan}
\author{Greg van Anders}
\email{gva@queensu.ca}
\affiliation{ 
Department of Physics\text{,} Engineering Physics and Astronomy, Queen's University\text{,} Kingston ON\text{,} Canada
}%

\date{\today}

\begin{abstract}

Data-driven causal relationship identification is pertinent to advancing understanding of complex systems both within and beyond science. Bayesian networks offer a probabilistic method for modelling generic causal relationships via directed acyclic graphs (DAGs). However, typical techniques for constructing Bayesian networks rely on optimization, which can be ill-suited for learning causal relationships because the underlying data may admit multiple chains of causation. More data-faithful representations of causal relationships would provide frameworks for constructing multiple causal maps that are consistent with the variability that is inherent in underlying data. Here, we show that entropy-based inference generates atlases of plausible causal relationships that are consistent with underlying data. 
On simulated noisy data of 2- and 20-node linear structural equation models, we sample a maximum-entropy ensemble of graphs that allow us to quantify the inherent structural ambiguity in underlying causal relationships. Our method shows that ``optimized'' DAGs can contain causal artifacts are not consistent across equivalently accurate topologies.

\end{abstract}


\maketitle

\section{Introduction}

Bayesian networks are probabilistic graphical models defined over directed acyclic graphs (DAGs), where directed edges encode conditional dependence relations among random variables. They provide a graphical representation of joint distributions and have applications in biology \citep{Sachs2005}, genetics \citep{Zhang2013}, causal inference \citep{spirtesCausation1993}, fairness and accountability \citep{Kusner2017}, and finance \citep{Sanford2012}. Learning a DAG that faithfully represents the data-generating distribution is challenging because the space of DAGs grows superexponentially with the number of variables, and because the acyclicity constraint induces a highly nonconvex feasible set \citep{zhengDAGs2018,belloDAGMA2023}. Moreover, observational data may not identify a unique graph: distinct DAGs can encode the same conditional independence relations, leading to Markov-equivalent structures that are equally consistent with the data \citep{Husmeier2005a}. As a result, returning a single optimized graph artificially binarizes potential causal relationships and can obscure the structural ambiguity that would faithfully represent underlying data.

Recent continuous formulations of DAG learning have made score-based structure learning substantially more scalable. NOTEARS introduced a smooth algebraic characterization of acyclicity, turning the combinatorial DAG constraint into a differentiable equality constraint suitable for first-order optimization \citep{zhengDAGs2018}. Subsequent methods extended this idea for alternative acyclicity characterization \citep{yuDAGGNN2019,zhengLearningSparseNonparametric2020,ngRoleSparsityDAG2020,yuDAGsNoCurl2021,belloDAGMA2023}. These algebraic characterizations identify a causal graph by first identifying the topology, and then estimating the conditional distributions or structural parameters. In practice, acyclicity-regularization based approaches (e.g., DAGMA, NOTEARS) converge from a generic graph to an acyclic one by reducing the influence of the underlying data over the course of training. It is therefore unclear the extent to which a score-based DAG actually represents causal relationships in underlying phenomena.

Bayesian DAG learning replaces point estimation with posterior inference over plausible graph structures. Instead of first choosing a topology and then fitting local conditional models, joint Bayesian formulations can represent uncertainty in both structure and parameters simultaneously. Existing methods, including VCN \citep{annadaniVariationalCausalNetworks2021}, DiBS \citep{lorchDiBSDifferentiableBayesian2021}, BayesDAG \citep{annadaniBayesDAGGradientBasedPosterior2023}, and ProDAG \citep{thompsonProDAGProjectedVariational2025}, approximate this posterior using different variational, sampling, regularization, or projection-based strategies. Although these methods differ, each imposes structural assumptions in their inference. For instance, ProDAG constructs distributions over sparse DAGs by first drawing unconstrained continuous matrices from an assumed probability distribution and then projecting them to the nearest weighted acyclic adjacency matrices inside an $\ell_1$-constrained region; the resulting projected distributions are used as both priors and variational posteriors \citep{thompsonProDAGProjectedVariational2025}. 
However, such formulations can bias the inferred posterior by encoding assumptions about sparsity, acyclicity, or projection geometry before inference begins. This suggests the need for an alternative formulation in which the prior-like structure is inferred from the data-supported landscape itself, rather than assumed a priori.

Here, we formulate Bayesian network structure learning from a maximum-entropy perspective. Rather than prescribing a structural prior over DAGs, we first construct a canonical ensemble over weighted graph parameters, allowing the relative importance of configurations to be determined by the score landscape and the volume of data-supported graphs. This ensemble induces an effective structural prior through its geometry. Exact acyclicity is imposed only after sampling, through a nonlinear projection to DAG space. The resulting acyclic ensemble preserves the statistical weights generated by maximum-entropy inference and provides edge-level and graph-level uncertainty beyond a single optimized DAG.

\section{Methodology}

\subsection{Notation and Background}

\paragraph{Notation.}
We use $[d]$ to denote the set $\{1,\dots,d\}$. The data matrix is denoted by $X$, and a directed graph on $d$ nodes is represented by a weighted adjacency matrix $W\in\mathbb{R}^{d\times d}$, where $W_{ij}$ corresponds to the directed edge $i\to j$. We write $\mathrm{pa}(j)$ for the set of parents of node $j$. A graph is a directed acyclic graph (DAG) if it contains no directed cycles. For matrices $A$ and $B$, $A\circ B$ denotes the Hadamard product. We denote by $\|\cdot\|_F$ the Frobenius norm, by $\|\cdot\|_p$ the vector $\ell_p$-norm, and by $\|\cdot\|_0$ the number of nonzero entries. Also, the notation $\nabla_W$ denotes the gradient with respect to $W$. 

\paragraph{Bayesian networks, graphs, and structural equation models.}
A Bayesian network is a probabilistic graphical model defined over a directed acyclic graph (DAG) $G = ([d], E)$, where each node $j \in [d]$ corresponds to a random variable $X_j$. The graph encodes conditional independence relationships among the variables and induces a factorization of the joint distribution as
\begin{equation}
P(X) = \prod_{j=1}^d P\big(X_j \mid X_{\mathrm{pa}(j)}\big),
\end{equation}
where $\mathrm{pa}(j)$ denotes the set of parents of node $j$ in $G$. This factorization implies that each variable is conditionally independent of its non-descendants given its parents. The DAG $G$ therefore provides a compact representation of the dependency structure among the variables. Learning a Bayesian network amounts to identifying this underlying graph from data.

An equivalent representation of this structure is given by structural equation models (SEMs). Let $X = (X_1, \dots, X_d)$ be a $d$-dimensional random vector. A (nonparametric) SEM represents each variable as
\begin{equation}
X_j = f_j(X, Z_j), \quad \forall j \in [d],
\end{equation}
where each $f_j : \mathbb{R}^{d+1} \to \mathbb{R}$ is a nonlinear function and $Z_j$ is an exogenous noise variable. We consider the Markovian setting, in which the variables $Z_j$ are mutually independent. Although each function $f_j$ depends only on a subset of variables, corresponding to the parents of $X_j$, we define each $f_j$ on the full space for notational simplicity. 

The collection $f = (f_1, \dots, f_d)$ induces both a joint distribution $P(X)$ and a graph $G(f)$, where edges correspond to functional dependencies. Under the independence assumption on the noise variables, the induced distribution factorizes according to $G(f)$, and thus defines a Bayesian network. Conversely, under mild conditions, any Bayesian network admits such a structural representation. Therefore, learning the structure of a Bayesian network is equivalent to identifying the dependency structure encoded by the functions $f_j$ in the SEM. This perspective allows us to study graph structure through functional representations, which will be central to our formulation.

\paragraph{Score-based structure learning.}
Given a joint distribution over $Z = (Z_1, \dots, Z_d)$, the SEM defines a distribution $P(X)$ and an associated graph $G(f)$. The goal is to recover $G(f)$ from $n$ i.i.d.\ samples of $X$. Let $X = [x_1, \dots, x_d] \in \mathbb{R}^{n \times d}$ denote the data matrix.

In score-based approaches, we define a score function $Q(f; X)$ to measure the quality of a candidate SEM:
\begin{equation}
Q(f; X) = \sum_{j=1}^d \mathrm{loss}(x_j, f_j(X)),
\end{equation}
where $f_j(X) \in \mathbb{R}^n$, and $\mathrm{loss}$ can be, for example, the least squares loss or a log-likelihood function. Given a function class $\mathcal{F}$, the structure learning problem is formulated as
\begin{equation}
\min_{f \in \mathcal{F}} Q(f; X) \quad \text{subject to} \quad G(f) \in \mathrm{DAGs}.
\end{equation}

\paragraph{Functional representation of graphs.}
Following prior work \cite{belloDAGMA2023}, we assume that each $f_j$ lies in a Sobolev space of square-integrable functions with square-integrable derivatives. Let $\partial_k f_j$ denote the partial derivative of $f_j$ with respect to $X_k$. Then $f_j$ is independent of $X_k$ if and only if $\|\partial_k f_j\|_{L_2} = 0$.

Using this observation, we define a weighted adjacency matrix $W(f) \in \mathbb{R}^{d \times d}$ with entries
\begin{equation}
[W(f)]_{i,j} := \|\partial_i f_j\|_{L_2},
\end{equation}
which encodes the graphical structure of the model. In particular,
\[
G(f) \in \mathrm{DAGs} \quad \Longleftrightarrow \quad W(f) \in \mathrm{DAGs},
\]
where $W(f)$ is interpreted as a weighted adjacency matrix.

In practice, the functions $f_j$ are chosen from a parametrized family (e.g., neural networks), making the problem finite-dimensional. The SEM formulation encompasses a wide range of models, including additive noise models, generalized linear models, additive models, polynomial regression, and index models. Here, we assume that the model class is chosen such that the underlying graph $G(W)$, and hence the corresponding Bayesian network, is identifiable from $W$. 

\subsection{Maximum Entropy Inference}

The functional representation above induces a weighted adjacency matrix $W(f)$ that encodes the dependency structure of the variables. In practice, it is convenient to work directly with this matrix representation and parameterize the model in terms of $W \in \mathbb{R}^{d \times d}$.

Under this parameterization, we consider a family of models in which the joint distribution of the data factorizes according to a DAG encoded by $W$. In particular,
\begin{equation}
p(X \mid W) = \prod_{j=1}^d p\big(x_j \mid X_{\mathrm{pa}_W(j)}\big),
\end{equation}
where $\mathrm{pa}_W(j)$ denotes the set of parents of node $j$ as specified by the support of $W$.

Rather than working directly with the likelihood, we consider a score function $Q(W; X)$ that measures the discrepancy between the model and the observed data. In general, $Q$ may include data fidelity terms as well as structural constraints, and should be interpreted as an effective energy or potential over graph structures.

A standard approach is to estimate $W$ by minimizing $Q(W; X)$ subject to $W \in \mathrm{DAGs}$. However, such optimization-based methods select a single solution and fail to capture the intrinsic ambiguity of the problem, since multiple graph structures may explain the data equally well.

To address this limitation, we instead consider a distribution over graph structures. Specifically, we seek a distribution $p(W \mid X)$ that maximizes entropy subject to constraints imposed by the data:
\begin{equation}
\max_{p(W \mid X)} 
\left[
- \int p(W \mid X)\, \log p(W \mid X)\, dW
\right],
\label{eqn:Sdef}
\end{equation}
subject to
\begin{equation}
\mathbb{E}_{p}[Q(W; X)] = \langle Q \rangle,
\qquad
\int p(W \mid X)\, dW = 1.
\label{eqn:Sconstr}
\end{equation}
Maximizing Eq.\ \eqref{eqn:Sdef} with respect to $p(W\mid X)$ subject to the constraints Eq.\ \eqref{eqn:Sconstr} gives the Gibbs distribution
\begin{equation}
p(W \mid X) = \frac{1}{Z(\beta)} \exp\big(-\beta \, Q(W; X)\big),
\label{eq:Wdist}
\end{equation}
where $\beta$ is the Lagrange multiplier that enforces the ``energy'' constraint and
\begin{equation}
\label{eq:partitionfunc}
Z(\beta) = \int \exp\big(-\beta \, Q(W; X)\big)\, dW,
\end{equation}
is a partition function that normalizes the distribution.

This defines a statistical ensemble over weighted graph parameters, where low-energy configurations are favored while alternative solutions are retained. The parameter $\beta$ controls the concentration of the distribution, interpolating between exploration and optimization.

At this stage, the ensemble is defined over unconstrained weighted graph parameters, so it should not yet be interpreted as a distribution supported only on DAGs. Exact acyclicity is enforced after sampling, through the nonlinear projection to DAG space described in Sec.~\ref{sec:AcycProj}. The final DAG ensemble therefore inherits its statistical weights from the maximum-entropy ensemble, while the projection step removes cyclic structure from each sampled graph.

\subsubsection{Implicit Bayesian inference over graph structures \label{sec:ImpBayesian}}
The Gibbs distribution in Eq.~\eqref{eq:Wdist} defines a canonical ensemble over weighted directed graph parameters. At this stage, $W$ is not required to be acyclic, and the ensemble should be viewed as a distribution over candidate weighted dependency structures rather than as a posterior supported on DAGs. The role of the score $Q(W;X)$ is to define the energy landscape explored by the ensemble: low-score graphs are favored, while finite temperature allows nearby and alternative low-score regions to contribute.

To characterize coarse-grained features of this ensemble, we introduce collective variables
\[
\boldsymbol{\theta}=\boldsymbol{\theta}(W,X),
\]
which summarize structural features induced by the sampled graph parameters. They need not be prescribed as optimization targets; rather, they are induced by the sampled weighted graphs and the data. By marginalizing the canonical ensemble over all graph parameters $W$ that give the same value of $\boldsymbol{\theta}$, we obtain 
\begin{equation}
Z(\beta)
=
\int d^{n_\theta}\theta
\int dW \;
e^{-\beta Q(W; X)}\,
\delta\!\big(\boldsymbol{\theta}(W, X) - \boldsymbol{\theta}\big),
\end{equation}
which can be expressed purely in terms of the collective variables as
\begin{equation}
Z(\beta)
=
\int d^{n_\theta}\theta\;
e^{-\beta F(\boldsymbol{\theta}, X)},
\end{equation}
where the effective free energy is defined implicitly by
\begin{equation}
e^{-\beta F(\boldsymbol{\theta}, X)}
=
\int dW \;
e^{-\beta Q(W; X)}\,
\delta\!\big(\boldsymbol{\theta}(W, X) - \boldsymbol{\theta}\big).
\end{equation}
This induces a distribution over the collective variables
\begin{equation}
p(\boldsymbol{\theta} \mid X)
\propto
e^{-\beta F(\boldsymbol{\theta}, X)}.
\end{equation}

Following the mean–value theorem, the exponential term in the definition of the free energy can be decomposed as
\begin{equation}
e^{-\beta F(\boldsymbol{\theta}, X)}
=
\big\langle e^{-\beta Q(W; X)} \big\rangle_{\boldsymbol{\theta}}
\;\Omega(\boldsymbol{\theta}, X),
\end{equation}
where $\langle \cdot \rangle_{\boldsymbol{\theta}}$ denotes an average over the hypersurface of $W$ yielding the same $\boldsymbol{\theta}$, and
\begin{equation}
\Omega(\boldsymbol{\theta}, X)
:=
\int dW \;
\delta\!\big(\boldsymbol{\theta}(W, X) - \boldsymbol{\theta}\big)
\end{equation}
is the corresponding volume of that hypersurface. Defining the entropy $S(\boldsymbol{\theta}, X) = \ln \Omega(\boldsymbol{\theta}, X)$, and assuming that $e^{-\beta Q(W; X)}$ varies slowly over each iso–$\boldsymbol{\theta}$ hypersurface, we obtain the free energy decomposition
\begin{equation}
F(\boldsymbol{\theta}, X)
=
\langle Q(W; X) \rangle_{\boldsymbol{\theta}}
-
\frac{1}{\beta} S(\boldsymbol{\theta}, X).
\end{equation}
Under the slow-variation approximation, in which fluctuations of $Q(W;X)$ over each iso-$\boldsymbol{\theta}$ hypersurface are neglected, the collective-variable distribution takes the form of an implicit Bayesian decomposition,

\begin{equation}
p(\boldsymbol{\theta} \mid X)
\;\propto\;
p(X \mid \boldsymbol{\theta}) \, \Omega(\boldsymbol{\theta}, X),
\end{equation}
where the likelihood is not specified explicitly, but is instead induced by the ensemble:
\begin{equation}
p(X \mid \boldsymbol{\theta})
\;\propto\;
\Big\langle e^{-\beta Q(W; X)} \Big\rangle_{\boldsymbol{\theta}}
\approx
\exp\!\big(-\beta \,\bar{Q}(\boldsymbol{\theta}, X)\big),
\end{equation}
with
\begin{equation}
\bar{Q}(\boldsymbol{\theta}, X)
=
\frac{1}{\Omega(\boldsymbol{\theta}, X)}
\int dW \;
Q(W; X)\,
\delta\!\big(\boldsymbol{\theta}(W, X) - \boldsymbol{\theta}\big).
\end{equation}
Here, $\exp(-\beta \bar Q(\boldsymbol{\theta},X))$ plays the role of an effective data-fit term, while $\Omega(\boldsymbol{\theta},X)$ acts as an entropic prior over structural collective variables. The free energy $F(\boldsymbol{\theta}, X)$ encodes a balance between data fidelity and structural robustness. The term $\langle Q(W; X) \rangle_{\boldsymbol{\theta}}$ favors collective variables for which there exist graph structures that fit the data well, while the entropic contribution $S(\boldsymbol{\theta}, X)=\ln \Omega(\boldsymbol{\theta}, X)$ rewards structural features that can be realized by many distinct graphs. 

This variability can be quantified empirically via the pairwise Structural Hamming Distance (SHD) between sampled graphs (see Sec.~\ref{sec:UncerQuant}). Regions of large volume $\Omega(\boldsymbol{\theta}, X)$ 
may be reflected empirically in graph diversity. High pairwise SHD indicates that the projected ensemble contains structurally distinct DAGs, whereas low pairwise SHD indicates concentration around a small set of graph structures.

In the limit $\beta \to \infty$, the entropy term vanishes and minimizing $F$ reduces to minimizing $Q$, recovering the classical optimization-based formulation. For finite $\beta$, however, entropy favors high-volume regions of graph space, biasing the solution toward structurally robust configurations and inducing variability across sampled graphs, which is reflected in increased pairwise SHD. In this sense, no explicit prior is required: the geometry of the model space induces an entropic prior $\Omega(\boldsymbol{\theta}, X)$, while the effective likelihood is governed by the average score.

\subsubsection{Molecular Dynamics Sampling \label{sec:MD}}

The Gibbs distribution
\begin{equation}
p(W \mid X) \propto \exp\!\big(-\beta Q(W; X)\big)
\end{equation}
defines a canonical ensemble over graph structures. To sample from this distribution, we use the Simmering method \citep{simmering}, a molecular-dynamics-based sampling method in which optimization parameters are treated as dynamical particles' positions evolving under an energy landscape defined by the objective function. In this framework, the entries of $W$ are interpreted as the positions of particles, while the score function $Q(W; X)$ acts as a potential energy. The negative gradient $-\nabla_W Q(W; X)$ therefore defines a force that drives the system toward regions of lower energy, analogous to gradient-based optimization. 

Rather than directly minimizing the score, Simmering \cite{simmering} introduces finite-temperature dynamics that allow the system to explore a canonical ensemble of graph configurations. In this setting, low-score graphs are favored, but higher-score graphs can also be visited with probabilities controlled by the inverse temperature parameter, $\beta$. This allows us to sample not only the single best graph, but a broader ensemble of causal structures. The main ingredients of the method are described below.

\paragraph{Augmented phase space.}
Following standard constructions \cite{simmering}, we introduce an auxiliary momentum variable for each parameter in $W$. Specifically, to each entry $W_{ij}$ we associate a conjugate momentum $P_{ij}$, and consider the augmented state $(W,P)$. This allows us to rewrite the partition function by multiplying it by unity in the form of a Gaussian integral over momenta,
\begin{equation}
1 = \frac{\int dP \, \exp\!\big(-\beta \tfrac{1}{2}\|P\|_F^2\big)}{Z_{\mathrm{free}}(\beta)},
\end{equation}
where $Z_{\mathrm{free}}(\beta)$ is the normalization constant of the free (kinetic) system. Inserting this identity into the partition function yields
\begin{equation}
Z(\beta)
=
\int dW \, dP \;
\exp\!\big(-\beta \mathcal{H}(W,P)\big),
\end{equation}
where the Hamiltonian is defined as
\begin{equation}
\mathcal{H}(W,P)
=
Q(W; X)
+
\frac{1}{2} \|P\|_F^2.
\end{equation}
This transformation leaves the partition function in Eq.~\eqref{eq:partitionfunc} unchanged, but lifts the problem from the configuration space of $W$ to an extended phase space $(W,P)$. 

The significance of this reformulation is that it enables a dynamical interpretation of the inference problem. While the original formulation defines a static distribution over graph structures, the Hamiltonian $\mathcal{H}(W,P)$ naturally generates equations of motion in phase space. These dynamics provide a practical way to generate finite-temperature trajectories whose stationary distribution is designed to match the target canonical distribution, allowing samples from $p(W \mid X)$ to be obtained by simulating trajectories rather than directly evaluating the high-dimensional integral.

To prevent the dynamics from collapsing to a single minimum, we couple the graph parameters to a Nos\'e-Hoover chain thermostat. The thermostat regulates the kinetic energy of the trajectory and allows the system to continue exploring multiple low-energy regions of the score landscape.

\paragraph{Nos\'e-Hoover chain dynamics.}
To generate samples at a fixed temperature $T = 1/\beta$, we employ a Nos\'e-Hoover chain (NHC) thermostat \citep{GlennMartyna1996,Tuckerman1992,frenkelsmita}. This method augments the system with a set of auxiliary variables that act as a heat bath, allowing energy exchange and generating trajectories designed to sample the canonical ensemble at the target temperature. In this formulation, the entries of the adjacency matrix $W$ are treated as the positions of real particles, and we associate to each $W_{ij}$ a velocity $v_{ij}$ and a mass $M_{ij}$.

The real particle dynamics are given by \citep{ReversibleNHT}
\begin{align}
    \dot{W}_{ij}(t) &= v_{ij}(t), \label{eqn:eom_W1} \\
    \dot{v}_{ij}(t) &= a_{ij}(t) - v_{s_1}(t)\, v_{ij}(t), \label{eqn:eom_W2}
\end{align}
where the $(i,j)$ indices denote quantities associated with the graph parameters, and the $s_k$ subscript denotes quantities associated with the $k$-th auxiliary (thermostat) particle.

The term $v_{s_1}(t)\, v    _{ij}(t)$ in Eq.\ \eqref{eqn:eom_W2} distinguishes this dynamics from standard gradient-based optimization. In particular, in momentum gradient descent this term would correspond to a constant damping coefficient. In contrast, $v_{s_1}(t)$ in NHC dynamics evolves in time and can take both positive and negative values, allowing the system to alternate between dissipative and energy-injecting regimes \citep{francaConformal2020}.

The auxiliary particle dynamics are given by
\begin{align}
    \dot{s}_k(t) &= v_{s_k}(t), \label{eqn:eom_s1} \\
    \dot{v}_{s_1}(t) &= a_{s_1}(t) - v_{s_2}(t)v_{s_1}(t), \label{eqn:eom_s2} \\
    \dot{v}_{s_k}(t) &= a_{s_k}(t) - v_{s_{k+1}}(t)v_{s_k}(t), \label{eqn:eom_s3}
\end{align}
for $k = 2, \dots, N_c$, with the convention $v_{s_{N_c+1}} = 0$.

The accelerations are defined as
\begin{align}
    a_{ij}(t) &= -\nabla_{W_{ij}} Q(W(t); X), \label{eqn:real_acc_W} \\
    a_{s_1}(t) &= \frac{1}{M_{s_1}}\!\left(\sum_{i,j} M_{ij} v_{ij}^2(t) - d^2 T_{\text{target}}\right), \label{eqn:virt_as1_W} \\
    a_{s_k}(t) &= \frac{1}{M_{s_k}}\!\left(M_{s_{k-1}} v_{s_{k-1}}^2(t) - T_{\text{target}}\right), \label{eqn:virt_ask_W}
\end{align}
where $M_{ij}$ and $M_{s_k}$ denote the masses of the real and auxiliary particles, respectively.

The acceleration of the real particles in Eq.\ \eqref{eqn:real_acc_W} is given by the negative gradient of the energy $Q(W;X)$, and therefore drives the system toward configurations that better explain the data. The coupling to the auxiliary variables prevents collapse to a single minimum and instead enables sampling from the canonical ensemble defined by $Q(W;X)$.

From a dynamical perspective, the system evolves under the combined influence of deterministic forces derived from $Q(W; X)$ and the energy exchange induced by the thermostat. At low temperatures, the dynamics concentrate near minima of $Q$, recovering optimization-based structure learning. At finite temperature, however, the system explores multiple basins of attraction, generating an ensemble of plausible graph structures. In this sense, inference is recast as a dynamical process in graph space, where the system ``simmers'' over the energy landscape. Rather than converging to a single solution, the dynamics maintain a distribution over structures that balances data fidelity and structural entropy, as captured by the free energy (see Sec.~\ref{sec:ImpBayesian}).

The sampling trajectory is generated by symplectic integration of the equations of motion in Eqs.~\eqref{eqn:eom_W1}--\eqref{eqn:eom_s3}. In this work, we use the Verlet algorithm, following Ref.~\cite{simmering}. After equilibration, weighted graph samples are collected from the finite-temperature trajectory. Each sampled graph is then passed through the nonlinear projection to DAG space described in Sec.~\ref{sec:AcycProj}. The resulting projected graphs form an empirical DAG ensemble whose probabilities are inherited from the sampling frequencies of the original canonical ensemble.

\subsection{Acyclicity Projection \label{sec:AcycProj}}
A central difficulty in learning Bayesian networks is enforcing the acyclicity constraint. In continuous formulations of structure learning, a common strategy is to replace the combinatorial constraint of acyclicity with a smooth function that vanishes if and only if the graph is a DAG.

When the acyclicity condition is treated as a regularization term rather than enforced as a hard constraint, the resulting energy or Pareto surface may contain both acyclic and cyclic graphs. Thus, although DAGs are included in the feasible low-energy regions, the sampled or optimized solutions are not guaranteed to be acyclic unless an additional projection or constraint-enforcement step is applied. In optimization-based approaches, one can iteratively rescale the score function so that it mainly selects the relevant basin, and then recover a single DAG within that region \cite{belloDAGMA2023}. However, this strategy still focuses on one optimized solution and may obscure other plausible causal structures represented by different DAGs in the same energy landscape. 

To address this limitation, we use the molecular-dynamics sampling method described in Sec.~\ref{sec:MD} to generate a canonical ensemble of possible graphs. We then project the sampled weighted graphs, which may contain cycles, onto candidate DAGs and construct an ensemble of acyclic graphs. For the projection step, we adopt the log-determinant acyclicity characterization used in DAGMA \citep{belloDAGMA2023}, since it is exact, differentiable, and well suited to our energy-based formulation. For each sampled graph, we perform a nonlinear projection to DAG space. This nonlinear projection follows the geometry of the acyclicity level sets rather than selecting a  DAG under an externally imposed distance or sparsity criterion. Repeating this procedure for all samples produces an ensemble of DAGs whose weights are inherited from the original canonical ensemble through their sampling frequencies. Thus, the method does not collapse the landscape to a single optimal graph, nor does it impose an additional sparsity-biased variational posterior; instead, it converts the maximum-entropy ensemble into a projected ensemble of plausible DAGs.

\paragraph{Acyclicity Characterization function} Among all acyclicity characterizations, the log-det regularizer\cite{belloDAGMA2023} does not diminish the contribution of long cycles, has larger and better-behaved gradients, and is empirically faster to evaluate in practice. Following Ref\cite{belloDAGMA2023}, we first define, for $s>0$, the domain
\begin{equation}
\mathcal{W}_s
:=
\left\{
W \in \mathbb{R}^{d\times d}
\;\middle|\;
\rho(W\circ W) < s
\right\},
\end{equation}
where $\rho(\cdot)$ denotes the spectral radius. On this domain, the matrix
\[
sI - W\circ W
\]
is an $M$-matrix, and hence has positive determinant and nonnegative inverse \cite{bermanNonnegativeMatricesMathematical1979}. This allows one to define the log-determinant acyclicity function
\begin{equation}
h^{s}_{\mathrm{ldet}}(W)
=
-\log\det\!\big(sI-W\circ W\big)+d\log s,
\end{equation}
and show that
\begin{equation}
h^{s}_{\mathrm{ldet}}(W)\ge 0,
\qquad
h^{s}_{\mathrm{ldet}}(W)=0
\;\Longleftrightarrow\;
W \in \mathrm{DAGs}.
\end{equation}
Thus, $h^{s}_{\mathrm{ldet}}$ is an exact acyclicity characterization.

The parameter $s$  determines the feasible domain $\mathcal{W}_s$: larger values of $s$ enlarge the domain, while smaller values shrink it. In particular, DAGs belong to $\mathcal{W}_s$ for every $s>0$, and the set $\mathcal{W}_s$ is path-connected. A useful property of $h^{s}_{\mathrm{ldet}}$ is that it admits a simple closed-form gradient,
\begin{equation}
\nabla h^{s}_{\mathrm{ldet}}(W)
=
2\big(sI-W\circ W\big)^{-T}\circ W.
\end{equation}
Moreover, its entries have a direct graph-theoretic interpretation: the $(i,j)$ entry of the gradient is nonzero if and only if the edge $i\to j$ participates in a cycle. Equivalently, the negative gradient $-\nabla h^{s}_{\mathrm{ldet}}(W)$ acts only on cyclic edges, shrinking them while leaving edges that do not belong to cycles unaffected. This selectivity makes $h^{s}_{\mathrm{ldet}}$ a natural choice for the nonlinear projection to DAG space.

\paragraph{Projection onto the DAG space}
The graphs sampled by the molecular-dynamics trajectory are not necessarily acyclic. Therefore, after sampling, we project each sampled graph onto the space of DAGs. In high-dimensional graphs, breaking all directed cycles usually requires coordinated changes to several edges. We therefore use a nonlinear projection based on gradient flow of the acyclicity function. Here, projection refers to a deterministic nonlinear flow toward the zero level set of the acyclicity function. 

Let $\widetilde W$ denote one sampled graph. Starting from $\widetilde W$, we minimize the log-determinant acyclicity function $h^{s}_{\mathrm{ldet}}(W)$ with respect to $W$, using a value of $s$ large enough so that the sampled graph lies in the domain $\mathcal W_s$. Since the gradient of $h^{s}_{\mathrm{ldet}}$ is normal to its level sets, this flow moves the sampled graph in the direction of steepest decrease of the acyclicity violation. Thus, the path bends through graph space and progressively removes cyclic structure. The projection is stopped when $h^{s}_{\mathrm{ldet}}(W(\tau))\leq \delta,$ for a small tolerance $\delta>0$. The resulting graph is then a projected DAG. If the thresholded graph is the all-zero graph, it is discarded, since it is formally acyclic but contains no causal structure. 

This defines a nonlinear projection map which maps each sampled graph $\widetilde W$ to a nontrivial projected DAG $D$. Because the original graphs $\widetilde W$ are sampled from a canonical ensemble, the resulting DAG ensemble is the push-forward of this canonical ensemble under the projection map. In other words, the probability of observing a projected DAG depends not only on its score, but also on the volume of sampled cyclic graphs that flow to it under the projection. This allows the final ensemble to retain information about multiple plausible causal structures rather than collapsing to a single optimized graph.

The inference pipeline is therefore as follows. We first define a score over weighted graph parameters and sample the corresponding maximum-entropy ensemble using molecular-dynamics trajectories. Each sampled weighted graph is then passed through the nonlinear acyclicity projection, thresholded, and retained only if it yields a nontrivial DAG. The final output is an empirical acyclic ensemble from which edge marginal probabilities, edge-weight variability, and graph-level diversity are estimated.

\subsection{Uncertainty quantification\label{sec:UncerQuant}}
A key advantage of our method is that it produces an ensemble of DAGs rather than a single graph estimate. Let
$$
\mathcal D_{\mathrm{ens}}
=
\left\{
D^{(1)},D^{(2)},\ldots,D^{(M)}
\right\}
$$
denote the final ensemble of nontrivial DAGs obtained after projection. We quantify structural uncertainty using the following quantities.

\begin{itemize}
    \item \textbf{Edge marginal probabilities:}
    For each ordered pair of nodes $(i,j)$, we estimate the probability that the directed edge $i\to j$ appears in the posterior ensemble by
    $$
    p_{ij}
    =
    \mathbb{P}(D_{ij}\neq 0\mid X)
    \approx
    \frac{1}{M}
    \sum_{m=1}^{M}
    \mathbf{1}\!\left\{
    D^{(m)}_{ij}\neq 0
    \right\}.
    $$
    Thus, $p_{ij}$ measures how frequently the edge $i\to j$ appears across the sampled DAGs. Edges with $p_{ij}$ close to one are strongly supported by the ensemble, while edges with intermediate values indicate structural ambiguity. Since our sampling procedure is not explicitly biased toward sparse graphs by an $\ell_1$ penalty, the projected DAGs may contain many numerically small but nonzero edge weights. Therefore, before computing uncertainty metrics, we threshold each weighted DAG. More specifically, for a fixed threshold $\epsilon>0$, an edge $i\to j$ is treated as present in the $m$-th sampled DAG if $|D_{ij}^{(m)}|>\epsilon$, and absent otherwise. In this way, very small edge weights are treated as numerical noise rather than meaningful causal relations.

    \item \textbf{Edge-weight variability:}
    In addition to whether an edge appears, we measure how much its weight varies across the ensemble. For each edge weight, we compute
    $$
    \mathrm{Var}(D_{ij})
    =
    \frac{1}{M}
    \sum_{m=1}^{M}
    \left(
    D^{(m)}_{ij}
    -
    \overline D_{ij}
    \right)^2,
    $$
    where
    $$
    \overline D_{ij}
    =
    \frac{1}{M}
    \sum_{m=1}^{M}
    D^{(m)}_{ij}.
    $$
    Large variance indicates that the strength of the relation between variables $i$ and $j$ is uncertain, even if the edge appears frequently.

\item \textbf{Graph diversity:}
    We measure diversity within the DAG ensemble using the average pairwise Structural Hamming Distance (SHD). Following the standard definition used in DAG structure learning, SHD counts the number of edge additions, deletions, and reversals needed to convert one graph into another \citep{belloDAGMA2023}. For two sampled DAGs $D^{(m)}$ and $D^{(n)}$, we write this distance as $\mathrm{SHD}\!\left(D^{(m)},D^{(n)}\right).
    $
    We then compute the average pairwise SHD across the ensemble,
    $$
    \overline{\mathrm{SHD}}
    =
    \frac{2}{M(M-1)}
    \sum_{1\le m<n\le M}
    \mathrm{SHD}
    \left(
    D^{(m)},D^{(n)}
    \right).
    $$
    A small value of $\overline{\mathrm{SHD}}$ means that most sampled DAGs have similar edge structures, indicating low structural uncertainty. A large value means that the sampled DAGs differ substantially from one another, indicating higher uncertainty in the inferred causal graph.

\end{itemize}

Together, these quantities summarize different aspects of uncertainty. Edge marginal probabilities measure support for individual causal relations, edge-weight variability measures uncertainty in the strength of those relations, and graph diversity measures uncertainty at the level of the whole graph. These ensemble-based diagnostics provide information about structural ambiguity and model confidence that is not available from point-estimate methods.

\section{Result}

We evaluate the proposed method in two complementary settings. First, we consider a low-dimensional illustrative example to visualize the geometry of the acyclicity constraint and the induced dynamics. Second, we study a synthetic linear structural equation model (SEMs) which is characteristic of standard benchmarks. 

\subsection{Illustrative example: noisy two-node system}

To visualize the geometry of the proposed energy landscape and the behavior of the sampling dynamics, we consider a two-node linear SEM with additive noise. Let the ground-truth graph be
\[
X_1 \longrightarrow X_2,
\]
and generate $n$ i.i.d.\ samples according to
\begin{align}
x_1^{(\ell)} &= \xi_1^{(\ell)}, \\
x_2^{(\ell)} &= a\,x_1^{(\ell)} + \xi_2^{(\ell)},
\qquad \ell = 1,\dots,n,
\end{align}
where $\xi_1^{(\ell)}$ and $\xi_2^{(\ell)}$ are independent zero-mean noise variables, and $a \sim \mathrm{Uniform}([-2,-0.5]\cup[0.5,2])$. In our experiment, $a\approx 0.9$, and we take $\xi_1^{(\ell)},\xi_2^{(\ell)}$ to be Gaussian,
\[
\xi_1^{(\ell)},\xi_2^{(\ell)} \sim \mathcal N(0,\sigma^2),
\]
with $\sigma=1$.
We denote the resulting data matrix by $X \in \mathbb R^{n\times 2}$.

We restrict the candidate weighted adjacency matrix to the form
\begin{equation}
W =
\begin{bmatrix}
0 & w_1\\
w_2 & 0
\end{bmatrix},
\end{equation}
so that $w_1$ corresponds to the edge $X_1 \to X_2$ and $w_2$ corresponds to the edge $X_2 \to X_1$. In this setting, acyclicity holds if and only if $w_1=0$ or $w_2=0$.

\paragraph{Data-fidelity term.}
We take the data-fidelity term to be
\begin{equation}
Q_{\mathrm{data}}(W)
=
\frac{1}{2n}
\|X - XW\|_F^2.
\end{equation}
For the present two-node model, this reduces to
\begin{equation}
Q_{\mathrm{data}}(w_1,w_2)
=
\frac{1}{2n}
\sum_{\ell=1}^n
\left[
\big(x_1^{(\ell)} - w_2 x_2^{(\ell)}\big)^2
+
\big(x_2^{(\ell)} - w_1 x_1^{(\ell)}\big)^2
\right].
\end{equation}
Defining the empirical second moments
\begin{equation}
c_{11} = \frac{1}{n}\sum_{\ell=1}^n \big(x_1^{(\ell)}\big)^2,
\qquad
c_{22} = \frac{1}{n}\sum_{\ell=1}^n \big(x_2^{(\ell)}\big)^2,
\qquad
c_{12} = \frac{1}{n}\sum_{\ell=1}^n x_1^{(\ell)}x_2^{(\ell)},
\end{equation}
we obtain
\begin{equation}
Q_{\mathrm{data}}(w_1,w_2)
=
\frac{1}{2}(c_{11}+c_{22})
- c_{12}(w_1+w_2)
+\frac{1}{2}\big(c_{11}w_1^2 + c_{22}w_2^2\big).
\end{equation}

Hence the gradient is
\begin{equation}
\nabla Q_{\mathrm{data}}(w_1,w_2)
=
\begin{bmatrix}
c_{11}w_1 - c_{12}\\
c_{22}w_2 - c_{12}
\end{bmatrix},
\end{equation}
and the Hessian is
\begin{equation}
\nabla^2 Q_{\mathrm{data}}(w_1,w_2)
=
\begin{bmatrix}
c_{11} & 0\\
0 & c_{22}
\end{bmatrix}.
\end{equation}
Thus, the data term alone is a strictly convex quadratic function.

\paragraph{Acyclicity projection}
We project the sampled $(w_1,w_2)$ pairs using the log-determinant acyclicity regularizer,
\begin{equation}
h_{\mathrm{ldet}}^{s}(W)
=
-\log\det(sI-W\circ W)+2\log s,
\end{equation}
with $s=1$. For the present two-node model,
\begin{equation}
W\circ W
=
\begin{bmatrix}
0 & w_1^2\\
w_2^2 & 0
\end{bmatrix},
\qquad
\det(sI-W\circ W)=s^2-w_1^2w_2^2,
\end{equation}
and therefore
\begin{equation}
h_{\mathrm{ldet}}^{s}(w_1,w_2)
=
-\log(s^2-w_1^2w_2^2)+2\log s.
\end{equation}
Its gradient is
\begin{equation}
\nabla h_{\mathrm{ldet}}^{s}(w_1,w_2)
=
\frac{1}{s^2-w_1^2w_2^2}
\begin{bmatrix}
2w_1w_2^2\\
2w_2w_1^2
\end{bmatrix},
\end{equation}

\paragraph{What to expect.}
This illustrative model provides a direct visualization of how sampled graphs flow to the nearest acyclic graph. The quadratic data-fidelity term $Q_{\mathrm{data}}$ admits a unique minimizer at $(w_1,w_2)=\big(c_{12}/c_{11},\,c_{12}/c_{22}\big)$, which typically lies in the interior where both edges are active, corresponding to a cyclic graph. The acyclicity regularizer, however, introduces a barrier that penalizes precisely such configurations, suppressing solutions with $w_1 w_2 \neq 0$.

As a result, the stationary points of the ``total energy''
\begin{equation}
    E(w_1,w_2)=Q(w_1,w_2)+h^s_{\mathrm{ldet}}(w_1,w_2)
\end{equation} include two acyclic configurations located along the coordinate axes:
\[
\left(\frac{c_{12}}{c_{11}},\,0\right),
\qquad
\left(0,\,\frac{c_{12}}{c_{22}}\right).
\]
These correspond to the two possible directed graphs $X_1 \to X_2$ and $X_2 \to X_1$. The acyclicity regularizer eliminates the interior cyclic minimizer of $Q_{\mathrm{data}}$ and instead splits the landscape into two competing basins centered around these axis-aligned stationary points.

Because the data are generated from the model $X_1 \to X_2$, the basin corresponding to $(w_1 \neq 0,\, w_2 \approx 0)$ attains lower total energy than its reverse counterpart. Consequently, at low temperature the Nos\'e--Hoover dynamics sample graphs that project to a region concentrated near the stationary point $\big(c_{12}/c_{11},\,0\big)$ associated with the data-aligned graph. At intermediate temperatures, the sampler may occasionally transition between the two basins, thereby revealing the uncertainty induced by finite sample size and observational noise. At sufficiently high temperature, the trajectories explore a broader region of the $Q_{\mathrm{data}}$ landscape, sampling graphs that flow to the neighborhoods of both basins and the intermediate acyclic graphs.

This experiment therefore serves three purposes. First, it illustrates the action of the acyclicity regularizer as a barrier that removes cyclic optima. Second, it illustrates how the total energy breaks the symmetry between competing directions by selecting between axis-aligned stationary points. Third, it demonstrates how finite-temperature sampling yields not only a preferred graph but also a distribution over plausible alternatives. 

\paragraph{Results.}

\begin{figure}
    \centering
    \includegraphics[width=\linewidth]{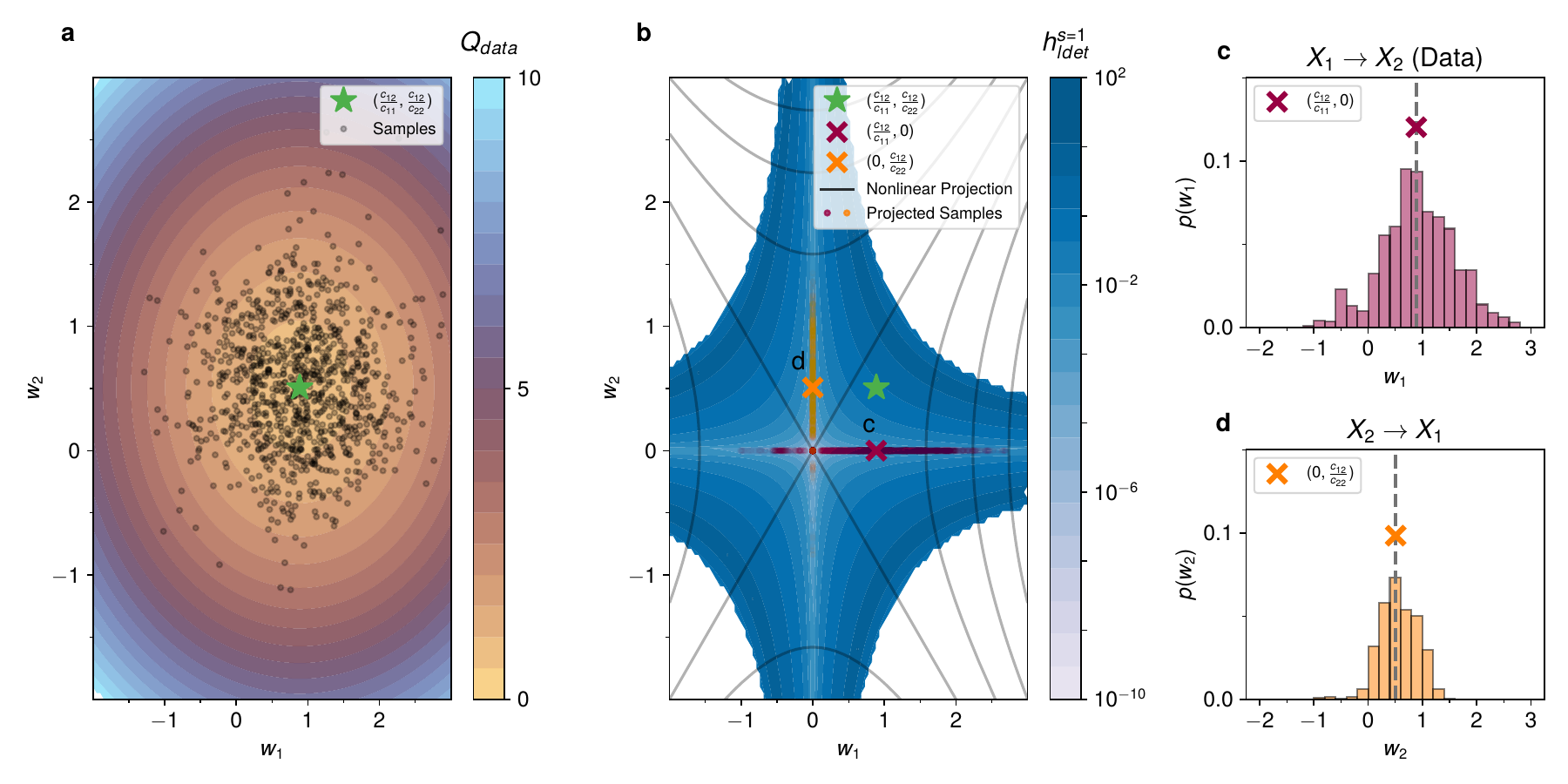}
    \caption{Finite-temperature sampling (at $T=0.5$) of the score loss landscape for the noisy 2-node system yields distributions over graph weights (panel c and d) corresponding to both possible directed acyclic graphs ($X_1 \rightarrow X_2$ and $X_2 \rightarrow X_1$ respectively), with a significantly higher proportion of samples for the data-consistent relationship. Our simulated data had a score minimizer of $(w_1,w_2)=(c_{12}/c_{11}, c_{12}/c_{22})\approx(0.89, 0.51)$. During the finite-temperature sampling stage, weight samples (round black markers, panel a) concentrate near the minimum of $Q_{\mathrm{data}}$ (green star marker, panel a) with some deviation proportional to the thermostat temperature. We equilibrated the thermostat for $1\times 10^6$ iterations with a timestep of $\Delta t=10^{-4}$, sampled $5\times10^6$ subsequent weights, and randomly selected $10^4$ of those weights for projection. After nonlinear projection on $h^{s=1}_\mathrm{ldet}$ (gradient level curves indicated with black lines, panel b), cyclic samples map onto the two axes in $w_1-w_2$ space corresponding to $h^{s=1}_\mathrm{ldet}=0$ (magenta and orange round markers, panel b), and the $Q_{\mathrm{data}}$ minimizer (green star marker, panel b) cyclicity is broken into two acyclic stationary points (magenta and orange $\times$ marker, panel b), each corresponding to one of the possible directed graphs ($X_1 \rightarrow X_2$ and $X_2 \rightarrow X_1$ respectively). After removing samples that map to the trivial solution ($(w_1,w_2)=0$), we find that the most probable projected weights correspond to the expected stationary points (peak of histogram in panels c,d vs. dashed line in panels c,d). }
    \label{fig:toy_model}
\end{figure}

Figure \ref{fig:toy_model} shows that the projected finite-temperature samples concentrate around regions of $h^s_\mathrm{ldet}$ that correspond to the analytical optima of the total energy $E(w_1,w_2)$, with more samples present along the axis corresponding to the data-aligned minimum (round magenta markers and magenta $\times$, Fig. \ref{fig:toy_model}b-c). Using Nos\'e-Hoover dynamics, we simmer on the score loss landscape and collect $(w_1,w_2)$ samples (round black markers, Fig. \ref{fig:toy_model}a), with samples closer to the minimizer $(w_1,w_2)=\big(c_{12}/c_{11},\,c_{12}/c_{22}\big)$ being energetically favourable. Comparing the loss landscapes and the $Q_\mathrm{data}$ minimizer (green star) in Fig. \ref{fig:toy_model}a and Fig. \ref{fig:toy_model}b, we see how the score loss minimizer gets projected onto the acyclic-graph generating regions of $h^s_\mathrm{ldet}$ (green star vs. orange and magenta $\times$, Fig. \ref{fig:toy_model}b) along the $\nabla h^s_\mathrm{ldet}$ level curves (black lines, Fig. \ref{fig:toy_model}b), turning the cyclic score loss optimum into two acyclic optima. The samples collected on the score loss landscape (Fig. \ref{fig:toy_model}a) flow into two modes along the two acyclic-graph producing axes (orange and magenta markers in Fig. \ref{fig:toy_model}b), each corresponding to a distribution centred on one of the acyclic optima. Using optimization, e.g., DAGMA \cite{belloDAGMA2023}, we would only recover the data-aligned minimizer of the total energy, and would have no insight into the inherent causal ambiguity of the problem. In contrast, our method both identifies the more probable graph (as shown by the larger proportion of samples projected into the $X_1 \rightarrow X_2$ relationship in Fig. \ref{fig:toy_model}c) and the plausible alternative causal relationship (orange-coloured sample flows onto the $w_1=0$ axis of $h^s_{\mathrm{ldet}}$ in Fig.  \ref{fig:toy_model}b, and distribution of weights in Fig. \ref{fig:toy_model}d), which would be inaccessible from a single, optimized graph.

\subsection{Linear SEM: Synthetic benchmark}

We evaluate our method on a synthetic linear structural equation model (SEMs), following the standard benchmark settings introduced in \cite{zhengDAGs2018} and later adopted by DAGMA (Appendix C.1) \cite{belloDAGMA2023}. This experiment provide a controlled environment for assessing both structural diversity and method scalability.

\paragraph{Data generation.}
We generate a random directed acyclic graph (DAG) with $d=20$ nodes using an Erd\H{o}s–R\'enyi (ER) graph with expected degree 4 (ER4). The expected degree is enforced by sampling approximately $4d$ edges and then randomly orienting them according to a topological ordering to ensure acyclicity.

Given a generated DAG with weighted adjacency matrix $W^\star \in \mathbb{R}^{d \times d}$, we sample edge weights independently from a uniform distribution
\[
W^\star_{ij} \sim \mathrm{Uniform}([-2,-0.5]\cup[0.5,2]),
\]
whenever there is an edge $i \to j$, and set $W^\star_{ij}=0$ otherwise. This avoids weak edges and ensures identifiability.

We generate a dataset of $n=1000$ i.i.d.\ samples according to the linear SEM
\begin{equation}
X = W^{\star\top} X + Z,
\end{equation}
or equivalently,
\[
X = (I - W^{\star\top})^{-1} Z.
\]
The noise matrix $Z \in \mathbb{R}^{n \times d}$ has independent entries sampled from Gaussian $\sim \mathcal{N}(0,1)$ distributions, allowing us to evaluate robustness to noise, as commonly done in prior work.

\paragraph{Model and score function.}
We define the data-fidelity term using the squared loss
\begin{equation}
Q_{\mathrm{data}}(W)
=
\frac{1}{2n}
\|X - XW\|_F^2,
\end{equation}
where $X \in \mathbb{R}^{n\times d}$ and $W \in \mathbb{R}^{d\times d}$.

Let
\[
x=\operatorname{vec}(X^T)\in\mathbb{R}^{nd},
\qquad
w=\operatorname{vec}(W^T)\in\mathbb{R}^{d^2},
\]
and define
\[
\Phi := I_d \otimes X \in \mathbb{R}^{nd\times d^2}.
\]
Using the identity
\[
\operatorname{vec}\big((XW)^T\big)=\operatorname{vec}(W^T X^T)=(I_d\otimes X)\operatorname{vec}(W^T),
\]
we obtain
\[
\operatorname{vec}\big((XW)^T\big)=\Phi w.
\]
Therefore, the data term can be written as
\begin{equation}
Q_{\mathrm{data}}(w)
=
\frac{1}{2n}
\|x-\Phi w\|_2^2.
\end{equation}

The gradient of the data term with respect to $w$ is
\begin{equation}
\nabla Q_{\mathrm{data}}(w)
=
\frac{1}{n}\,\Phi^T(\Phi w-x)
\in\mathbb{R}^{d^2\times 1},
\end{equation}
and its Hessian is
\begin{equation}
\nabla^2 Q_{\mathrm{data}}(w)
=
\frac{1}{n}\,\Phi^T\Phi
=
\frac{1}{n}\,(I_d\otimes X^T X)
\in\mathbb{R}^{d^2\times d^2}.
\end{equation}

As in the illustrative model, the ``total energy'', combining the influence of the score loss and the acyclic regularizer after projection, is defined as
\begin{equation}
E(W;X)
=
Q_{\mathrm{data}}(W)
+
h_{\mathrm{ldet}}^{s}(W).
\end{equation} 

\paragraph{Expected behavior.}
In this benchmark, classical methods aim to identify a single optimal DAG by minimizing a score function under acyclicity constraints (i.e., minimizing $E(W; Q)$ directly). In contrast, our method samples from a distribution over graph structures. At low temperatures, the sampler concentrates near global minima of $Q(W;X)$, recovering solutions comparable to optimization-based methods. At intermediate temperatures, the sampler explores multiple high-probability DAGs, capturing structural uncertainty due to finite samples and noise.

\paragraph{Results.}

\begin{figure}
    \centering
    \includegraphics[width=\linewidth]{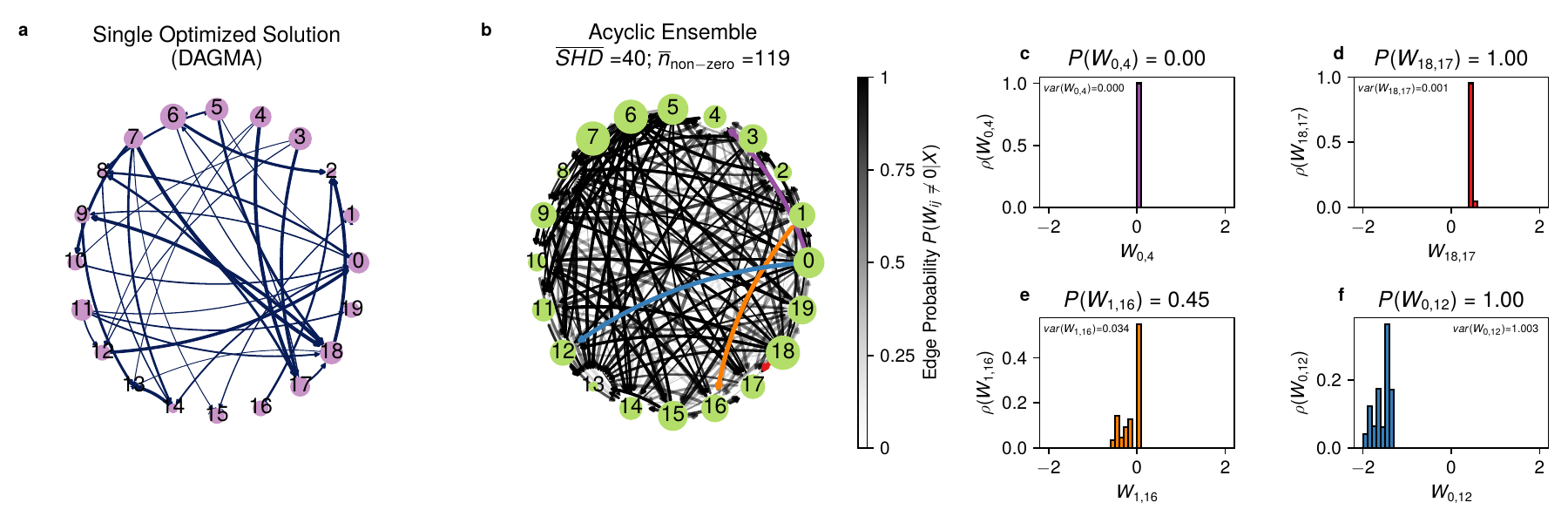}
    \caption{Finite-temperature sampling at $T=1$ yields a diverse ensemble of graphs that offers insight into structural uncertainty in a 20-node ER4 problem. Optimization using DAGMA \cite{belloDAGMA2023} (default fitting parameters) (panel a) yields a single graph with a single set of weights (thickness of blue edges, panel a), which prohibits investigation into the confidence of identified causal relationships. The acyclic ensemble was equilibrated for $1.2\times 10^5$ timesteps with a timestep $\Delta t=10^{-4}$, after which $4\times 10^4$ samples were collected, $1\times 10^4$ of which were randomly chosen to be projected. The sampled acyclic ensemble has a SHD of 40, which, relative to the average number of non-zero nodes (119), indicates high topological diversity among sampled graphs. This structural diversity is also reflected in the edge probability distribution (edge opacity and colourbar, panel b). From the set of possible edges, four were chosen (colour-coded between panel b and c--f) to display four trends in marginal edge probability (title, panels c--f) and edge variance (top left corner, panels c--f). Panel c shows a weight which has low probability of being present across the ensemble, and low variance, suggesting that its absence is robust. Panel e shows a weight which also has low probability but higher variance, indicating that the data constrains its absence less strongly. Analogous to panel c, panel d shows a weight with high marginal probability and low variance, indicating robustness of presence and of relationship strength. In contrast, panel f shows a weight with high probability and high variance, suggesting that the causal relationship between the corresponding nodes is more ambiguous. Note that in panel a and b, the size of the nodes indicates the degree of the optimized graph and marginal edge probability graph,  respectively. }
    \label{fig:linear_sem}
\end{figure}

Figure \ref{fig:linear_sem} shows that our sampler explores a distribution of probable DAGs, and offers direct observation and characterization of structural uncertainty via edge probabilities (edge opacity in Fig. \ref{fig:linear_sem}b) and weight distributions (shown for selected edges from Fig. \ref{fig:linear_sem}b in Fig. \ref{fig:linear_sem}c--f). Optimizing the total energy directly yields only one graph with one set of connection strengths (represented by edge thickness in Fig.\ref{fig:linear_sem}a, generated by using the DAGMA methodology \cite{belloDAGMA2023} on the problem setup described above). Using our sampler, we gain access to multiple uncertainty quantification measures without compromising the acyclicity requirement. For this problem, under finite-temperature sampling, we can identify a diverse set of distinct graph structures, as shown by the large average SHD between ensemble members ($\overline{SHD}=40$, Fig. \ref{fig:linear_sem}b) relative to the average number of non-zero edges ($\overline{n}_{\mathrm{non-zero}}=119$). From this ensemble of graphs, we can quantify the edge marginal probabilities (edge opacity in Fig. \ref{fig:linear_sem}b), which reflect how frequently particular causal relationships manifest across the ensemble. At a per-edge level, we can also characterize how strongly the data constrains the causal relationship strength between nodes:  edge magnitude variance is shown as edge thickness in Fig. \ref{fig:linear_sem}b, and the full distribution of sampled magnitudes for four edges are shown in Fig. \ref{fig:linear_sem}c--f. We can categorize the trends of edge distributions into four groups according to the concurrence of marginal edge probability and edge magnitude variance. Figure \ref{fig:linear_sem}c shows an edge which has low variance and low marginal edge probability, i.e., it is absent in all members of the ensemble. Figure \ref{fig:linear_sem}e also shows an edge that is likely to be absent, but with a higher marginal edge probability and higher variance (some ensemble members have small negative weights for this edge, and some have a zero-weight edge). Both edges shown in Fig. \ref{fig:linear_sem}c,e are absent from the single, optimized graph in Figure \ref{fig:linear_sem}a, but their absence is not equally guaranteed in the ensemble, highlighting the importance of this ensemble analysis in robust causal relationship characterization. Figure \ref{fig:linear_sem}d shows an edge with high marginal probability and low variance, which is also present in Fig. \ref{fig:linear_sem}a, indicating a strong relationship between those nodes in the data generation process. In contrast, Fig. \ref{fig:linear_sem}f shows a highly probable edge with higher variance, whose reverse is present in Fig. \ref{fig:linear_sem}a, suggesting that high edge variance may be also related to causal direction ambiguity in causal learning.

To understand this concretely, our analysis in Fig.\ \ref{fig:linear_sem}f indicates that the edge $12\to0$ in the DAGMA-produced DAG is potentially an artifact of constrained optimization that may not be supported by the underlying data. Our analysis uncovered consistent DAGs in which the direction of causality runs in the opposite direction. This example illustrates that finite-temperature graph ensemble sampling offers valuable insight into causal relationships that cannot be characterized from direct optimization alone. 

\section{Discussion}
This work proposes a probabilistic learning strategy for directed acyclic graphs based on maximum-entropy inference. Rather than prescribing a prior distribution over DAGs and then performing posterior inference using a likelihood induced by a chosen score function, we allow an effective structural prior to emerge from the geometry of the score landscape itself. From this perspective, graph configurations that occupy larger data-supported regions of parameter space receive greater statistical weight, while configurations with lower score are still favored by the energy landscape. The resulting ensemble therefore represents both data fidelity and structural uncertainty: edge frequencies quantify how consistently a causal relation appears across the projected DAG ensemble, while edge-weight variability captures uncertainty in the strength of those relations. More broadly, the method does not return a single optimized graph, but an ensemble of plausible DAGs, allowing multiple data-compatible causal structures to be represented simultaneously.

In realistic settings, causal structure is rarely reducible to a single identifiable “true” DAG. Hidden variables, measurement effects, and proxy relations can induce dependencies among observed variables, mask conditional independencies, or make several graph structures statistically indistinguishable under observational data. Moreover, in continuous DAG-learning formulations, the entries of a weighted adjacency matrix should not always be interpreted as phenomenological causal strengths: they often serve as relaxed parameters that encode the support of a binary graph, unless the structural equation model gives them a direct physical or statistical meaning. This makes two-stage procedures conceptually limited. Even if the local conditional distributions are fitted accurately after a topology is chosen, the selected topology already imposes a particular causal description of the data. Bayesian DAG methods address this issue by placing uncertainty over graph structures, but their posterior distributions still depend on assumed graph priors, variational families, ordering variables, sparsity penalties, or projection rules. Our approach follows a different principle. Rather than specifying the structural prior in advance, we let an effective prior emerge from the maximum-entropy ensemble itself. The result is not a single causal graph, but an atlas of data-supported acyclic structures, in which edge frequencies, weight variability, and graph diversity describe how causal relations are distributed across the ensemble.

 This viewpoint is natural from the perspective of statistical physics. In physical systems, macroscopic behavior is rarely explained by selecting one microscopic configuration as the correct state of nature. Instead, observable structure emerges from an ensemble of configurations, shaped by both energy and entropy. A phase, an order parameter, or a collective variable is meaningful not because it is imposed a priori, but because many microscopic states organize around it. We adopt the same logic for Bayesian network learning. Rather than assuming that the data must be summarized by one privileged DAG, we treat candidate graph structures as configurations in a statistical ensemble. The score landscape plays the role of an energy, while the volume of compatible graph configurations supplies the entropic contribution. Causal relations are therefore interpreted as emergent collective features of the ensemble: an edge is important not simply because it appears in one optimized graph, but because it persists across many data-supported acyclic configurations. In this sense, the learned network is not a single frozen diagram of causality, but a thermodynamic portrait of causal atlases.

\bibliography{bib_BN}
\end{document}